\documentclass[letterpaper]{article} 
\usepackage[draft]{aaai25}  
\usepackage{times}  
\usepackage{helvet}  
\usepackage{courier}  
\usepackage[hyphens]{url}  
\usepackage{graphicx} 
\usepackage{natbib}  
\usepackage{caption} 
\usepackage{algorithm}  
\usepackage{algorithmic}
\usepackage{amsmath}
\usepackage{amssymb}
\usepackage{booktabs}
\usepackage{multirow}
\usepackage{pifont}
\usepackage{newfloat}
\usepackage{listings}
\DeclareCaptionStyle{ruled}{labelfont=normalfont,labelsep=colon,strut=off} 
\floatstyle{ruled}
\newfloat{listing}{tb}{lst}{}
\floatname{listing}{Listing}
\title{AM-SAM: Automated Prompting and Mask Calibration for Segment Anything Model}
\author{
    Yuchen Li,
    Li Zhang,
    Youwei Liang,
    Pengtao Xie
}
\affiliations{
    Department of Electrical and Computer Engineering, University of California, San Diego, USA\\
    yul272@ucsd.edu, include.zl.h@gmail.com, youwei@ucsd.edu, p1xie@ucsd.edu
}

\nocopyright

\usepackage{bibentry}

\begin{document}
\maketitle
\begin{abstract}
Segment Anything Model (SAM) has gained significant recognition in the field of semantic segmentation due to its versatile capabilities and impressive performance. Despite its success, SAM faces two primary limitations: (1) it relies heavily on meticulous human-provided prompts like key points, bounding boxes or text messages, which is labor-intensive; (2) the mask decoder's feature representation is sometimes inaccurate, as it solely employs dot product operations at the end of mask decoder, which inadequately captures the necessary correlations for precise segmentation. Current solutions to these problems such as fine-tuning SAM often require retraining a large number of parameters, which needs huge amount of time and computing resources. To address these limitations, we propose an automated prompting and mask calibration method called AM-SAM based on a bi-level optimization framework. Our approach automatically generates prompts for an input image, eliminating the need for human involvement with a good performance in early training epochs, achieving faster convergence. Additionally, we freeze the main part of SAM, and modify the mask decoder with Low-Rank Adaptation (LoRA), enhancing the mask decoder's feature representation by incorporating advanced techniques that go beyond simple dot product operations to more accurately capture and utilize feature correlations. Our experimental results demonstrate that AM-SAM achieves significantly accurate segmentation, matching or exceeding the effectiveness of human-generated and default prompts. Notably, on the body segmentation dataset, our method yields a 5\% higher dice score with a 4-example few-shot training set compared to the SOTA method, underscoring its superiority in semantic segmentation tasks. 
\end{abstract}

%

\section{Introduction}\label{sec:1}
Semantic segmentation, which is a fundamental task in computer vision, involves partitioning an image into semantically meaningful regions, with each pixel assigned a semantic label corresponding to the object or region it represents. This task is critical for a wide range of applications, including autonomous driving~\cite{xu2017end,cheng2020panoptic,rossolini2023real}, medical image analysis~\cite{ronneberger2015u,chen2021transunet}, robotic vision~\cite{tzelepi2021semantic} and augmented reality~\cite{zhang2020slimmer}. Over the years, numerous approaches have been developed to tackle semantic segmentation, leveraging advances in deep learning and convolutional neural networks (CNNs)~\cite{lecun1998gradient}. Traditional methods for semantic segmentation, such as Fully Convolutional Networks (FCNs)~\cite{long2015fully}, U-Net~\cite{ronneberger2015u}, and SegNet~\cite{badrinarayanan2017segnet}, have demonstrated significant success by utilizing end-to-end training to learn pixel-wise classification. 
Despite their effectiveness, these methods often require extensive training on large labeled datasets, which can be time-consuming and computationally expensive. 

With the advent of foundation models~\cite{bommasani2022opportunitiesrisksfoundationmodels}, also known as large pre-trained models, which are popular in Natural Language Processing, make the landscape of computer vision tasks evolve significantly at the same time. These models leverage vast amounts of data and computational resources to pre-train models on diverse datasets, which can then be fine-tuned for specific tasks with comparatively less effort and data. Recently, SAM (Segment Anything Model)~\cite{kirillov2023segment}, a prominent development in this field, has shown remarkable capabilities in semantic segmentation across various natural image datasets. SAM's versatility and high-performance of segmenting objects and generating masks as output stem from relevant human input, by providing prompts such as points, bounding boxes and text messages.

However, SAM's superior performance of segmentation faces significant limitations due to its reliance on these meticulous human-provided prompts, such as points and boxes information. This is inconvenient in many cases. For example, in real-time road scenes, which need at least one boxes as prompt input, the users need to provide several boxes to make it segment better. Moreover, SAM also faces the challenge of unsatisfying performance on particular downstream tasks in special fields, such as medical images segmentation. This is due to the huge distribution discrepancy between medical images and  natural images, which are used for training SAM. In many cases, despite appropriate prompts provided by humans, SAM cannot make good segmentation of these medical images~\cite{zhang2023customizedsegmentmodelmedical}. Additionally, the feature representation in SAM's mask decoder is often inadequate, as it relies solely on dot product operations. This approach fails to capture the complex correlations needed for precise segmentation, resulting in less accurate feature representations and, consequently, sub-optimal segmentation outcomes.
\begin{figure}[t]
\centering
\includegraphics[width=\columnwidth]{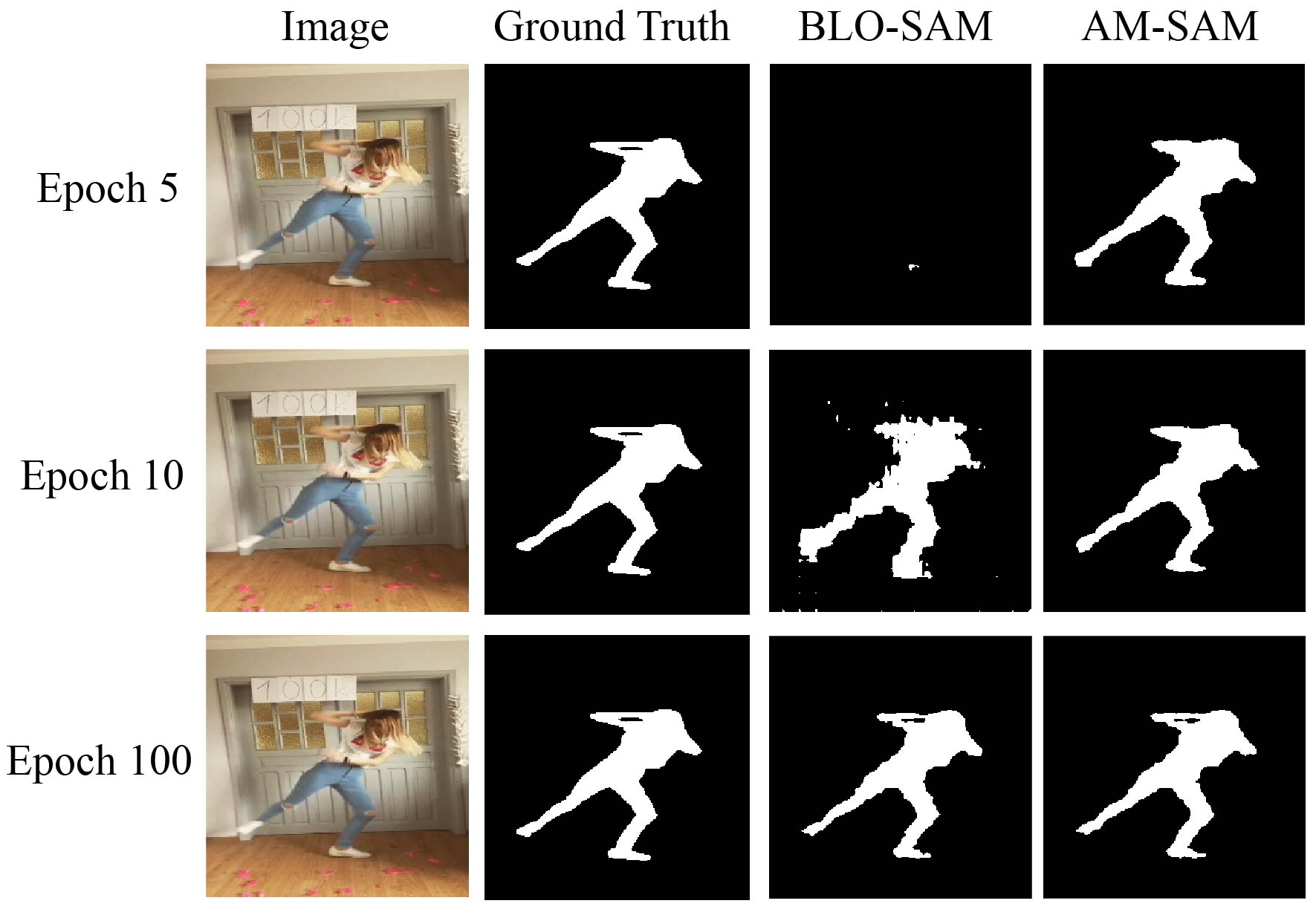} 
\caption{Masks generated during different training epochs. We trained both of the models in a few-shot training set with only 4 examples. AM-SAM has much better performance during the early training epochs with a faster converged model.}
\label{motivation}
\end{figure}

To address limitations of the distribution discrepancy and the need for manual prompt, SAMed~\cite{zhang2023customizedsegmentmodelmedical} applies the low-rank-based (LoRA) finetuning strategy with a default prompt, which is a learnable embedding to replace the prompts provided by human, and outperforms SAM in medical image datasets. However, SAMed suffers from a serious overfitting problem due to lack of segmentation masks of datasets in medical domains. To mitigate this issue, BLO-SAM~\cite{zhang2024blosambileveloptimizationbased} proposes a bi-level optimization method for the learnable prompt vector, demonstrating significant performance across datasets from different domains. Despite the good performance of BLO-SAM, it leverages the optimization method by initializing the prompt embedding with random values, similar to SAM's default prompt. This initial prompt value lacks explicit meaning and fails to represent prompt information, such as points and boxes. Consequently, during the early phase of the model's training process, there is a lack of accurate and proper position information, causing the poor score and inaccurate generated masks as shown in Figure~\ref{motivation}, which means the model needs more epochs to converge.

To address this weakness, we propose an automated prompting and mask calibration method called AM-SAM (\textbf{A}utomated prompting and \textbf{M}ask calibration for \textbf{SAM}), based on a bi-level optimization framework (BLO-SAM)~\cite{zhang2024blosambileveloptimizationbased}, which is illustrated in Figure~\ref{fig2}. It integrates an object detector to get accurate bounding box information, which enables the model to automate the generation of prompts, thus eliminating the need for human involvement and significantly reducing the manual effort required. By using the object detector, our method can accurately identify objects within the image, providing precise bounding boxes that serve as initial prompts for the segmentation task.

Furthermore, we enhance the mask decoder's feature representation by incorporating an element-wise multiplication in addition to the traditional dot product. This enhancement can be seen as a form of mask calibration for feature fusion, allowing our method to capture more complex feature correlations. The resulting masks are then combined with weights, effectively utilizing these enhanced feature correlations and leading to improved segmentation outcomes.

We also conducted extensive experiments on three datasets on general or medical domains to validate the effectiveness of our method. The results demonstrate that our approach not only achieves segmentation accuracy comparable to human-generated prompts but also surpasses the performance of existing methods on these datasets. This underscores the exceptional performance and effectiveness of our method in semantic segmentation tasks.

Our proposed AM-SAM framework systematically improves the generated prompts and segmentation masks, making the three main contributions:
\begin{itemize}
    \item \textbf{Automated Prompting and Initial Optimization}: We utilize an object detector to automatically generate bounding boxes that serve as initial prompts, which significantly enhances the model's segmentation performance and accelerates convergence during training.
    \item \textbf{Mask Calibration}: We modify the mask decoder of SAM which processes these prompts, propose a mask calibration method by incorporating advanced feature correlation techniques such as element-wise multiplication.
    \item \textbf{Effective Performance}: Our method demonstrates superior performance across both general and medical domains. It achieves a dice score which is 5\% higher than the SOTA method with a 4 example few-shot dataset on body segmentation dataset. We also have over 10\% higher performance than baseline method on medical domain dataset when trained with an 8 examples few-shot training set.
\end{itemize}






\section{Related Work}

\subsection{Semantic Segmentation}
Semantic segmentation is a fundamental task in computer vision that involves partitioning an image into semantically meaningful regions, where each pixel is assigned a label corresponding to the object or region it represents. Over the years, numerous approaches have been developed to tackle semantic segmentation, leveraging advances in deep learning and convolutional neural networks (CNNs). Fully Convolutional Networks (FCNs)~\cite{long2015fully} replace fully connected layers with convolutional ones to maintain spatial resolution. Subsequent advancements include encoder-decoder architectures like U-Net~\cite{ronneberger2015u} and SegNet~\cite{badrinarayanan2017segnet}, which employ downsampling followed by upsampling to capture multi-scale context while preserving spatial details. Attention mechanisms and Transformer-based architectures, such as the Vision Transformer (ViT)~\cite{dosovitskiy2020image} and Swin Transformer~\cite{liu2021swin}, have been applied to semantic segmentation. 

Recently, large-scale pre-trained models have begun to make significant impacts across various computer vision tasks, including semantic segmentation. These models, which are pre-trained on massive datasets and can be fine-tuned for specific tasks, have demonstrated remarkable versatility and effectiveness. For instance, foundation models such as CLIP~\cite{radford2021learning} have been adapted for segmentation tasks by leveraging their robust feature representations. These models enable strong generalization capabilities, especially when transferred to domain-specific applications. Building on this trend, SAM (Segment Anything Model)~\cite{kirillov2023segment} represents a significant advancement, designed to segment any object in an image, making it particularly suited for a wide range of segmentation tasks.
\begin{figure*}[t]
\centering
\includegraphics[width=1.0\textwidth]{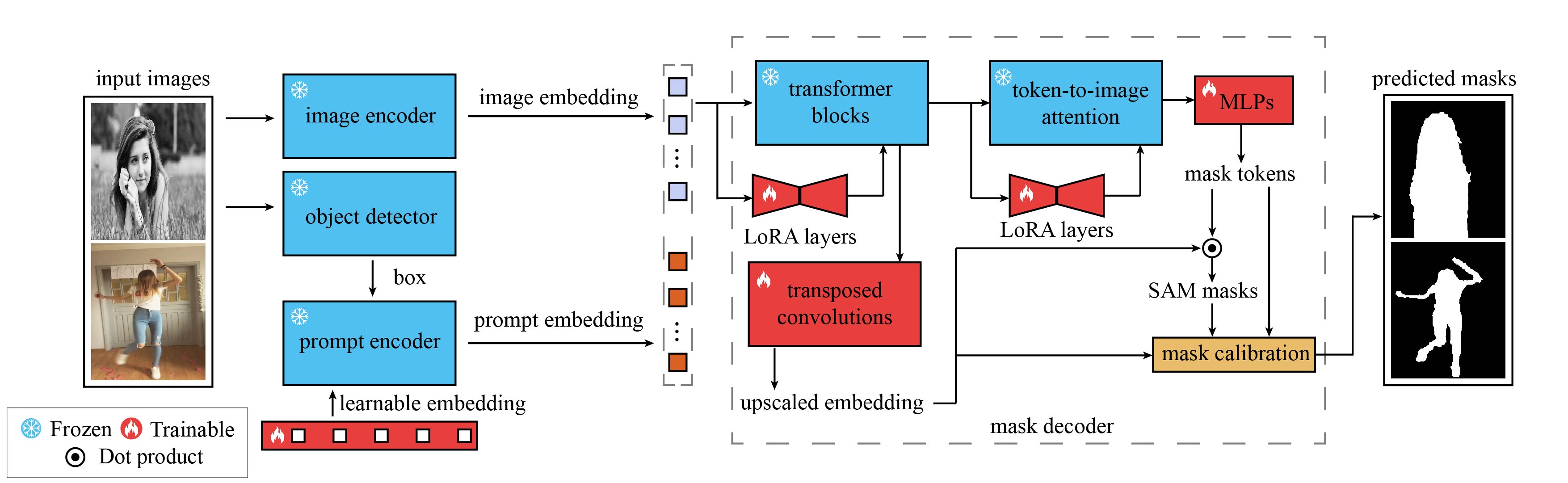} 
\caption{Overview of AM-SAM: an object detector is inserted as prompt generation network to get accurate position information and faster converging speed of the model. We keep it and the main parts of SAM parameters frozen. The mask decoder of SAM is modified by adding a mask calibration part to improve the performance of predicted masks.}
\label{fig2}
\end{figure*}
\subsection{Segment Anything Model (SAM)}
Just as mentioned, one of the most notable advancements in semantic segmentation is SAM, which has demonstrated remarkable capabilities across various natural image datasets by effectively segmenting objects and generating masks with human-provided prompts such as points, bounding boxes, and text messages. SAM employs a Vision Transformer (ViT)~\cite{dosovitskiy2020image} pre-trained with Masked Autoencoders (MAE)~\cite{he2022masked} to encode the input image. Positional embeddings are used to incorporate prompt information, and a lightweight Transformer-based mask decoder generates high-quality segmentation masks. To solve the defect of time cost and need for manual operation because of SAM's reliance on these meticulous prompts, Fast SAM~\cite{zhao2023fast} introduced YOLO~\cite{redmon2016you} series models to SAM and retrained the whole original SAM parameters. SAMed~\cite{zhang2023customizedsegmentmodelmedical} finetuned SAM with LoRA(Low-Rank Adaptation)~\cite{hu2021lora} by directing using the default prompt embedding SAM, and AutoSAM~\cite{shaharabany2023autosam} designed a prompts generator network to train these prompt embeddings on medical images, which are helpful in the automated points or boxes prompting progress. Further enhancing SAM, BLO-SAM~\cite{zhang2024blosambileveloptimizationbased} introduces a bi-level optimization approach with a few-shot training set, demonstrating significant improvements in segmentation accuracy with limited training data. EVF-SAM~\cite{zhang2024evf} exploits multi-modal prompts and comprises a pre-trained vision-language model to generate referring prompts. Recently, SAM 2~\cite{ravi2024sam} is released to support the segmentation of both videos and images based upon SAM. Anyway, SAM' has much room for improvement due to its dependence on human-provided prompts and poor performance on medical scenarios.

\section{Preliminaries}
\subsection{Low-Rank Adaptation (LoRA)}
Fine-tuning large pre-trained models has become an essential technique for adapting these models to specific downstream tasks. One of the most efficient and effective fine-tuning strategies is Low-Rank Adaptation (LoRA)~\cite{hu2021lora}, which focuses on optimizing a minimal set of parameters while leveraging the vast pre-trained knowledge within the model. In this work, we use LoRA to finetune the mask decoder of SAM inspired by SAMed~\cite{zhang2023customizedsegmentmodelmedical}. This method introduces trainable low-rank decomposition matrices into the existing layers of a model, which allows for efficient adaptation to new tasks without the need to modify the entire model. In the transformer blocks, we particularly focus on the self-attention mechanism. In a standard transformer block, the self-attention mechanism computes the attention scores using the query ($Q$), key ($K$), and value ($V$) matrices:
\begin{equation}
\text{Attention}(Q, K, V) = \text{Softmax}\left(\frac{Q K^T}{\sqrt{d_k}}\right) {V},
\end{equation}
where $Q = W_QX$, $K = W_KX$, and $V = W_VX$. Here, $X \in \mathbb{R}^{B \times N \times C_{in}} $ represents the input tokens, and $W_Q \in \mathbb{R}^{C_{in} \times C_{out}}, W_K \in \mathbb{R}^{C_{in} \times C_{out}}$, and $W_V \in \mathbb{R}^{C_{in} \times C_{out}}$ are the frozen weight matrices of SAM, where $C_{in}, C_{out}$ are the input and output channels, respectively. To apply LoRA, we decompose the weight update matrix $\Delta W$ into two low-rank matrices $A\in \mathbb{R}^{r \times C_{in}}$ and $B\in \mathbb{R}^{C_{out} \times r }$, where $r$ represents the rank of them, satisfying $r \ll \min \{ C_{in}, C_{out} \}$:
\begin{equation}
\Delta W = BA.
\end{equation}
Thus, the updated weight matrix $\hat{W}$ can be expressed as:
\begin{equation}
\hat{W} = W + \Delta W = W + B A.
\end{equation}
When applying this to the self-attention mechanism, we modify the projection matrices for the query and value matrices. The modified self-attention mechanism can be written as:
\begin{equation}
\begin{aligned}
\hat{Q} &= \hat{W}_Q X= W_QX + B_Q A_QX, \\
\hat{K} &= W_K X, \\
\hat{V} &= \hat{W}_V X= W_VX + B_V A_VX.
\end{aligned}
\end{equation}

\subsection{Bi-Level Optimization}
In this section, the bi-level optimization method in SAM implemented by BLO-SAM~\cite{zhang2024blosambileveloptimizationbased} will be explained, which is also used in our work.  In the lower level, they set the prompt embedding $A$ fixed, and split our training set into two halves to form the sub-dataset $D_1$ and $D_2$. The loss function used for optimizing the trainable parameters $W$ in SAM, including the parameters and weights of LoRA, MLPs and transposed transformers can be written as:
\begin{equation}\label{eq1}
\mathcal{L} = (1 - \lambda) \mathcal{L}_{CE}(W, A; D_1) + \lambda \mathcal{L}_{Dice}(W, A; D_1),
\end{equation}
where $\mathcal{L}_{CE}$ and $\mathcal{L}_{Dice}$ denote the cross-entropy loss, respectively, calculated between the predicted masks and the ground truth masks. Cross-entropy loss measures the discrepancy between them and dice loss evaluates the overlap between them, offering robustness against class imbalance. The parameter $\lambda$ serves as a trade-off factor. This stage focuses on determining the optimal model parameters $W$ by minimizing the loss on the split dataset $D_1$, given the current state of the prompt embedding $A$. The optimization problem is defined as:
\begin{equation}
W^*(A) = \arg\min_W \mathcal{L}(W, A; D_1).
\end{equation}
In the upper level, the primary goal is to optimize the learnable prompt embedding $A$ such that the loss on another dataset $D_2$, for validation, is minimized. This can be mathematically represented as:
\begin{equation}
\min_{A} \mathcal{L}(W^*(A), A; D_2).
\end{equation}
Therefore, the bi-level optimization framework can be represented as: 
\begin{equation}
\begin{aligned}
    &\min_{A} \mathcal{L}(W^*(A), A; D_2) \\
    &\text{s.t.} \quad W^*(A) = \arg\min_{W} \mathcal{L}(W, A; D_1).
\end{aligned}
\end{equation}
This bi-level optimization ensures that the prompt embeddings and model parameters are jointly optimized, leveraging different datasets for each optimization level. The upper-level optimization refines the prompt embeddings based on the performance of the model with parameters optimized in the lower level, facilitating a synergistic training process that enhances the model's segmentation capabilities.
\section{Method}
\subsection{Overview}
In our work, as shown in Figure~\ref{fig2}, AM-SAM first pre-processes the input images, then feeds the images into both the image encoder to get image embeddings and the object detector to generate bounding box as box prompts, respectively. After that, we perform a bi-level optimization step proposed by BLO-SAM~\cite{zhang2024blosambileveloptimizationbased} to train the parameters. Note that we also add a mask calibration part in the mask decoder to improve the performance of output masks. Moreover, we make it an end-to-end process, which means that we only need to provide the images once as input, and then get the segmented masks as outputs with improved performance.
\subsection{Automated Prompting}
In this part, we choose YOLOv8~\cite{yolov8_ultralytics} as the object detector, and integrate this YOLOv8 object detection model with SAM by inserting it as the the prompt generation network, keeping all the parameters of YOLOv8 frozen. After pre-processing the input images, we take them as input to the YOLOv8 model to generate bounding box prompts from the detected objects. During this process, we select the bounding box with the highest confidence score. From the selected bounding box, we extract the coordinates 
$(x_1, y_1)$ and $(x_2, y_2)$, which represent the top-left and bottom-right corners of the bounding box, respectively. The coordinates are then formatted into a tensor suitable as the input of the prompt encoder in SAM for further processing. 

A significant enhancement in our approach is the introduction of an initial prompt embedding generated by YOLOv8. This initial prompt embedding provides accurate and proper position information right from the start, which greatly improves the early training accuracy. Consequently, during the early phase of the model's training process, the availability of accurate position information reduces the number of epochs required for the validation score to converge. This is in stark contrast to previous methods where the lack of proper position information in the initial phase caused slower convergence and required more epochs to achieve optimal performance.

By leveraging this initial prompt embedding, our method ensures faster convergence with just a few training epochs, thereby enhancing the overall efficiency of the training process. This improvement is particularly advantageous in few-shot learning scenarios where training data is limited, as it allows the model to quickly adapt and achieve high accuracy with fewer training examples.
\begin{figure*}[t]
\centering
\includegraphics[width=1.0\textwidth]{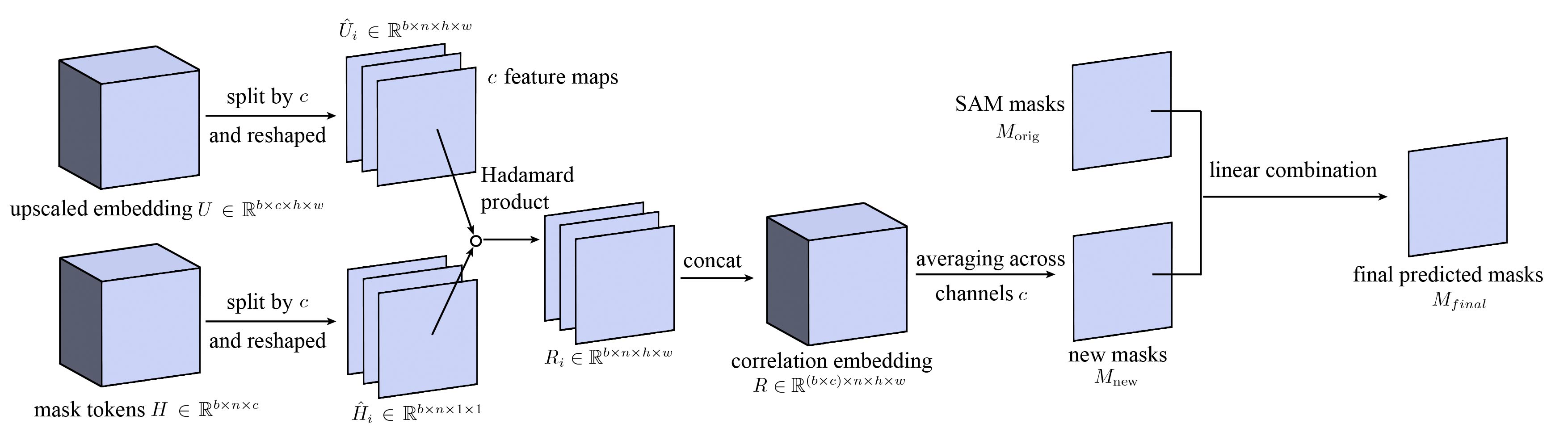} 
\caption{Steps of mask calibration of AM-SAM. We split the upscaled embedding and mask tokens, perform element-wise mutiplication and concatenate them for averaging and fusion. Then we combine it with SAM's original masks as final prediction.}
\label{mask_calibration}
\end{figure*}
\subsection{Mask Calibration}
In our proposed method, as shown in Figure~\ref{mask_calibration}, we introduce a mask calibration process to enhance the accuracy of predicted segmentation masks. At the end of the mask decoder of SAM, the trainable transposed convolution networks produce an upscaled image embedding, denoted as $U\in \mathbb{R}^{b \times c \times h \times w}$, where $b, c, h, w$ represent the batch size, number of channels, the height and width of the feature maps respectively; and the MLPs, with updated output token embedding in SAM's mask decoder~\cite{kirillov2023segment} as input, produce a transformed representation of mask tokens, denoted as $H\in \mathbb{R}^{b \times n \times c}$, where $n$ represents the number of mask outputs, aligning with the number of mask tokens. Then we split $U$ into $c$ embeddings, called $U_i\in \mathbb{R}^{b \times 1 \times h \times w}$, and $H$ into $c$ embeddings, called $H_i\in\mathbb{R}^{b \times n \times 1} $. For each corresponding $U_i$ and $H_i$, we repeat $U_i$ in the second dimension to expand it as $\hat{U}_i\in \mathbb{R}^{b \times n \times h \times w}$, and reshape $H_i$ as $\hat{H}_i\in \mathbb{R}^{b \times n \times 1 \times 1}$. To calibrate the masks, we perform element-wise multiplication, which is also known as Hadamard Product, denoting the operator as $\circ$, to get $c$ embeddings $R_i\in \mathbb{R}^{b \times n \times h \times w}$:
\begin{equation}
R_i=\hat{U}_i \circ \hat{H}_i.
\end{equation}
After that, we get a fused embedding called $R\in \mathbb{R}^{(b\times c) \times n \times h \times w}$ by concatenating them together: 
\begin{equation}
R=\text{Concat}(R_1, R_2, \ldots,R_c).
\end{equation}
This fused embedding better represents the correlations of features. By getting the average across the first dimension of $R$, we get $\Bar{M}\in \mathbb{R}^{1 \times n \times h \times w}$:
\begin{equation}
\Bar{M} = \frac{1}{b\times c} \sum_{i=1}^{b\times c} R_{i,n,h,w}.
\end{equation}
Then we repeat it in the first dimension to expand it to match the dimension of the original predicted masks $M_\text{orig}$, to get the new masks, denoted as $M_{new}\in \mathbb{R}^{b \times n \times h \times w}$.

The final step of our mask calibration process involves a weighted combination of the original predicted masks $M_\text{orig}$ and the newly generated masks:
\begin{equation}\label{eq12}
M_{final}=\alpha \cdot M_\text{orig} + (1 - \alpha) \cdot M_\text{new}.
\end{equation}
Here, $\alpha$ is a predefined weighting factor, controlling the balance between the original masks and the new mask. This procedure is essential to leverage the strengths of both the original and refined mask predictions, thereby enhancing the overall segmentation accuracy. The original masks provide a reliable baseline, while the new masks contribute enhanced accuracy through refined feature representations. The result is a more robust and precise segmentation output, which is particularly beneficial in applications requiring high accuracy, such as medical imaging and autonomous driving.

\section{Experiments}
In this section, we evaluate our AM-SAM in three datasets of semantic segmentation tasks. Our experiments are conducted in a few-shot learning setting, specifically focusing on scenarios with fewer than ten training examples per class.
\subsection{Datasets}
For our experiments, we utilized three datasets. The first one is the Segmentation Full Body TikTok Dancing Dataset~\cite{anwar2021tiktok} which consists of 2,615 images of segmented individuals engaged in various dance routines. 
To evaluate our model, we divided the dataset into a training set and a test set same as mentioned in BLO-SAM. The second one is Vikram Shenoy Human Segmentation Dataset~\cite{shenoy_human_segmentation_2024}. The Vikram Shenoy Human Segmentation Dataset available on GitHub is a collection of images focused on segmenting human figures in various environments. It includes 300 images of humans with some background and a corresponding binary mask for each of these images. We split the last 100 examples as the test set. The last one is ISIC 2018 dataset~\cite{codella2018skin,tschandl2018ham10000}, which is a prominent dataset in the field of medical image analysis, particularly in the study of dermatology for skin lesion segmentation and classification.
\subsection{Experimental Settings}
\begin{table}[t]
\centering
\caption{Test Dice Score (\%) on Tiktok Human Body Segmentation.}
\begin{tabular}{lcc} 
\toprule
Method & 4 Examples & 8 Examples \\ 
\midrule
DeepLab   & 31.8 & 37.0 \\ 
SwinUnet  & 30.9 & 56.8 \\ 
\cline{1-3}
HSNet     & 50.6 & 55.8 \\ 
SSP       & 58.9 & 76.5 \\ 
\cline{1-3}
SAM       & 26.0 & 26.0 \\ 
Med-SA    & 58.8 & 80.6 \\ 
SAMed     & 63.8 & 81.5 \\ 
\cline{1-3}
BLO-SAM   & 76.3 & 84.0 \\ 
\textbf{AM-SAM}  & \textbf{81.3} & \textbf{86.7} \\ 
\bottomrule
\end{tabular}
\label{tab:tiktok_human_body_4_8_examples}
\end{table}

\begin{table}[t]
\centering
\caption{Test Dice Score (\%) on Vikram Shenoy Human Segmentation Dataset.}
\begin{tabular}{lccc}
\toprule

Method & 4 Examples & 8 Examples & 16 Examples \\ 
\midrule
BLO-SAM & 89.1 & 91.3 & 92.4 \\ 
\textbf{AM-SAM} & \textbf{90.4} & \textbf{92.9} & \textbf{93.3} \\ 
\bottomrule
\end{tabular}
\label{tab:githubdataset_4_8_examples}
\end{table}

\begin{table}[t]
\centering
\caption{Test Dice Score (\%) on ISIC 2018 Dataset.}
\begin{tabular}{lccc}
\toprule
Method & 4 Examples & 8 Examples & 16 Examples \\ 
\midrule
BLO-SAM  & 65.1 & 62.3 & 71.0 \\ 
\textbf{AM-SAM}  & \textbf{66.3} & \textbf{73.6} & \textbf{71.9}\\ 
\bottomrule
\end{tabular}
\label{tab:ISIC2018_8_16_examples}
\end{table}
\begin{figure}[t]
\centering
\includegraphics[width=0.9\columnwidth]{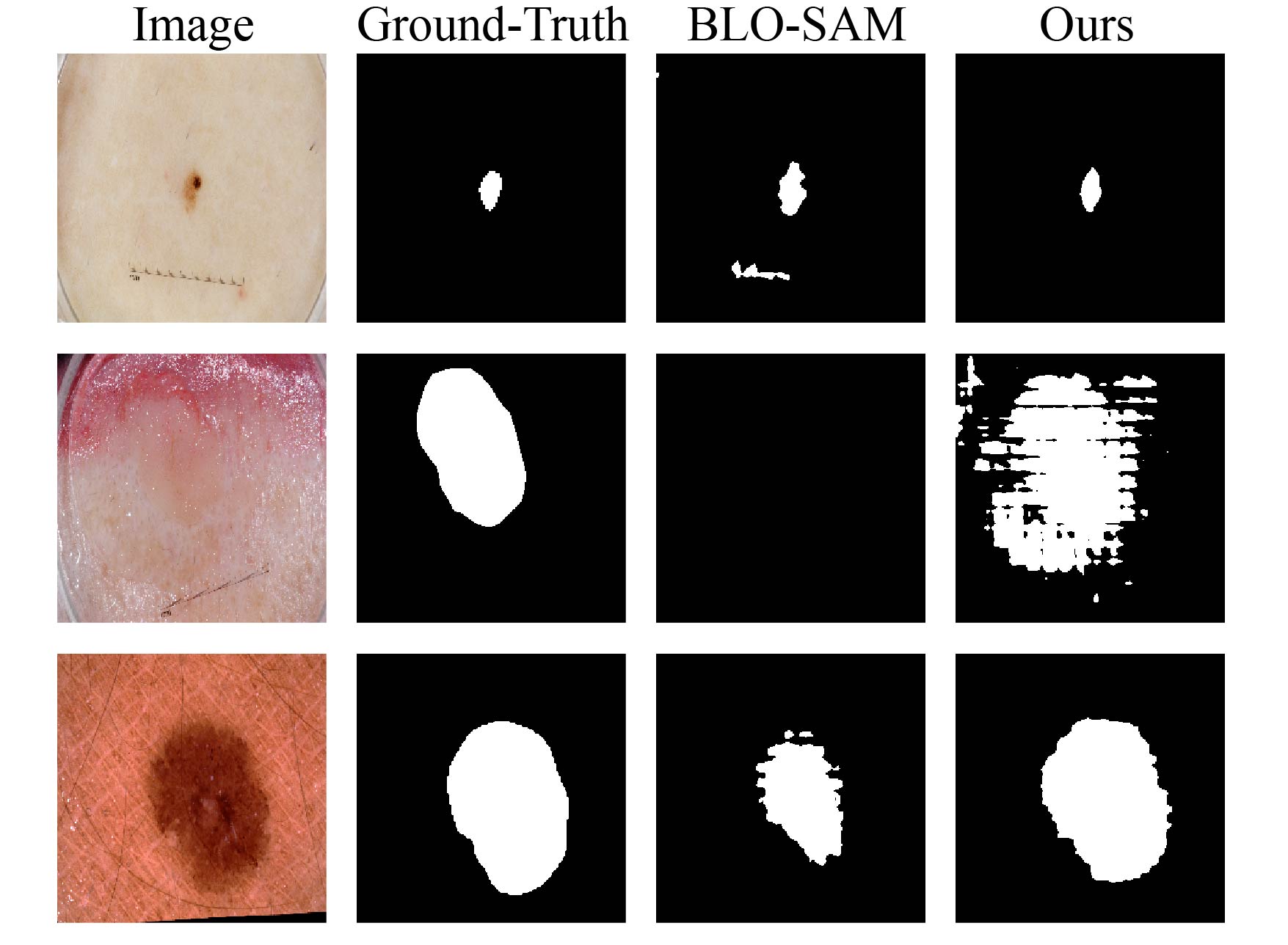} 
\caption{Qualitative results on some randomly sampled test examples from ISIC dataset.}
\label{ISIC}
\end{figure}
\subsubsection{Baselines and Metrics}
We compared our method with a variety of baselines, including supervised, few-shot, and SAM-based approaches. The supervised baselines comprise DeepLabV3~\cite{chen2017deeplab} and SwinUnet~\cite{cao2022swin}. Few-shot learning methods include HSNet~\cite{zhang2022hsnet} and SSP~\cite{fan2022self}. SAM-based baselines include vanilla SAM~\cite{kirillov2023segment}, Med-SA~\cite{wu2023medical} and SAMed~\cite{zhang2023customizedsegmentmodelmedical}, which used LoRA~\cite{hu2021lora} for finetuning. The most important method we want to compare it with is BLO-SAM~\cite{zhang2024blosambileveloptimizationbased}, since we implement the automated prompting and mask calibration based on its bi-level optimization framework, and improve its performance.
All the experiments are conducted on a NVIDIA GeForce RTX 3090 GPU with 24 GB of GDDR6X memory.

Our study utilizes the Dice score as the primary metric for evaluation. This metric quantitatively measures the similarity between the predicted mask $A$ and the ground truth mask 
$B$. The formulation of the Dice score is detailed below:
\begin{equation}
\text{Dice score} = \frac{2 \cdot |A \cap B|}{|A| + |B|}.
\end{equation}
\subsubsection{Hyper-parameters}
\begin{table*}[t]
\centering
\caption{Ablation studies on the YOLOv8 based automate prompting part based and the Hadamard Product based mask calibration part.}
\begin{tabular}{lcccccc}
\toprule
\multirow{2}{*}{Method} & \multirow{2}{*}{YOLO Detector} & \multirow{2}{*}{Mask Calibration} & \multicolumn{2}{c}{ISIC} & \multicolumn{2}{c}{TikTok Body}\\ 
\cmidrule(lr){4-5} \cmidrule(lr){6-7}
& & & \textbf{4 examples} & \textbf{8 examples} & \textbf{4 examples} & \textbf{8 examples}\\
\midrule
BLO-SAM  & \ding{55} & \ding{55} & 65.1 & 62.3 & 76.3 & 84.0 \\ 
AM-SAM  & \ding{55}& \ding{51}& 65.8 & 68.6 & 82.4  & 84.8 \\ 
AM-SAM  &\ding{51} &\ding{55} & \textbf{67.7} & 71.8 & 82.7 & 85.2 \\ 
\textbf{AM-SAM}  &\ding{51} & \ding{51}& 66.3 & \textbf{73.6}& \textbf{83.4} & \textbf{86.7} \\ 
\bottomrule
\end{tabular}
\label{tab:ablation}
\end{table*}
In our method, the trade-off parameter $\lambda$ in Equation~\ref{eq1} was maintained at 0.8, consistent with the settings from SAMed~\cite{zhang2023customizedsegmentmodelmedical}. The weighting factor $\alpha$ in Equation~\ref{eq12} in the mask decoder is set as 0.7. The optimization of LoRA layers and other non-frozen components was carried out in the lower-level using the AdamW optimizer~\cite{loshchilov2017decoupled}, with an initial learning rate of 5e-3, betas of $(0.9, 0.999)$, and a weight decay of 0.1. The learning rate for the $i$-th iteration was adapted using the formula from SAMed~\cite{zhang2023customizedsegmentmodelmedical}: $lr_i = lr_0 \left(1 - i/Iter_{max}\right)^{0.9}$, where $lr_0$ and $Iter_{max}$ denote the initial learning rate and the maximum number of iterations, respectively. 

For the upper-level optimization, we modified the initial learning rate to 1e-3 while keeping all other settings identical to those used in the lower-level optimization. The number of training epochs was set to 100, and the best checkpoint was selected based on segmentation performance on the $D_2$ sub-dataset.
\subsection{Results and Analysis}
In this section, we mainly use BLO-SAM~\cite{zhang2024blosambileveloptimizationbased} as baseline method. We first do the experiment in Full Body TikTok Dancing Dataset~\cite{anwar2021tiktok}, as shown in Table~\ref{tab:tiktok_human_body_4_8_examples}.
AM-SAM achieves significantly better performance with only 4 examples, with a score of 81.3\%, compared to the previous state-of-the-art method, BLO-SAM, which scores 76.3\%. This result strongly suggests that AM-SAM has superior capability in learning from a very small number of samples, which is a crucial advantage in practical applications where labeled data is limited. We also do the experiment on Vikram Shenoy Human Segmentation Dataset, the outcome is shown in Table~\ref{tab:githubdataset_4_8_examples}, which helps prove that our method is applicable on multiple datasets. The results on ISIC 2018 dataset, which is shown in Table~\ref{tab:ISIC2018_8_16_examples}, prove that AM-SAM also has good performance in medical scenarios. The visualization of qualitative results is shown in Figure~\ref{ISIC}. Since in most cases the use of center points of these boxes to create point prompts improves little, in this experiment we only generate box prompts to save training time. 

\begin{figure}[t]
\centering
\includegraphics[width=0.9\columnwidth]{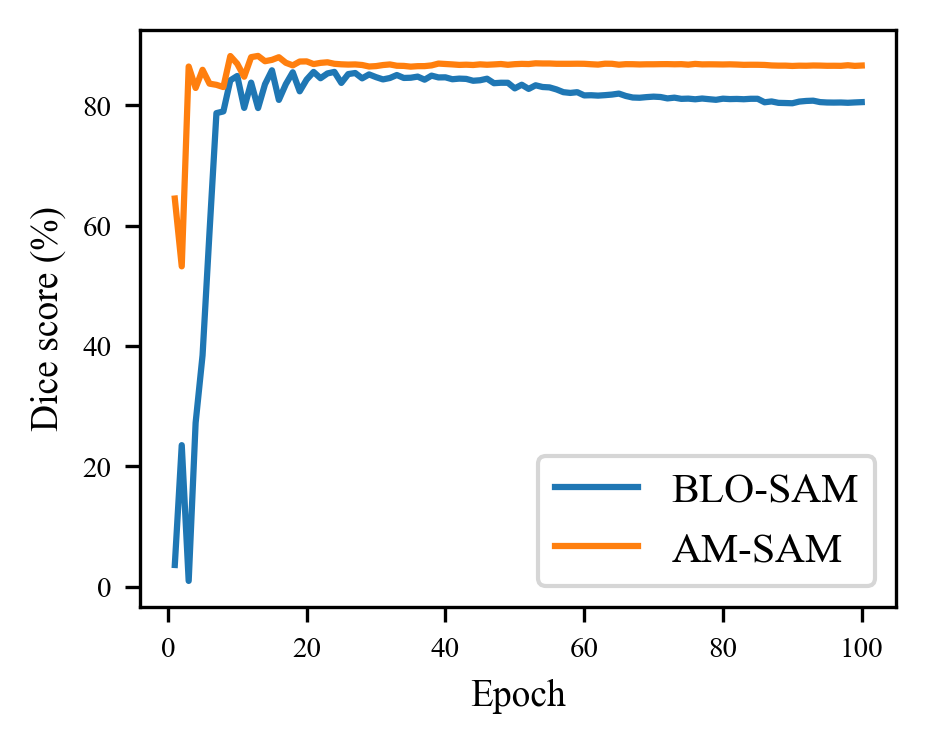} 
\caption{Comparison of validation dice score between BLO-SAM and AM-SAM when trained with only 8 examples. Notably the performance at the early stage of our AM-SAM is much better than that of BLO-SAM.}
\label{fig3}
\end{figure}

In the initial phase of training, our method exhibits a remarkable ability to converge rapidly, particularly when using a small number of training samples. As illustrated in Figure~\ref{fig3}, when training with only 8 examples, our model achieves a significant level of performance after just 5 epochs. Moreover, the model begins to stabilize and converge by the 10th epoch. The visualization can also be found in Figure~\ref{motivation}. This rapid convergence not only highlights the efficiency of our method but also underscores its superiority in few-shot learning scenarios. Compared to baseline methods such as BLO-SAM, which require more epochs to reach similar levels of performance, our approach demonstrates a clear advantage in terms of training speed and resource efficiency. The provided figure clearly illustrates this, showing that our method outperforms BLO-SAM both in the early stages of training and in the overall convergence process. This efficiency is particularly beneficial in scenarios where training data is scarce and fast convergence is critical.

\begin{figure}[t]
\centering
\includegraphics[width=0.9\columnwidth]{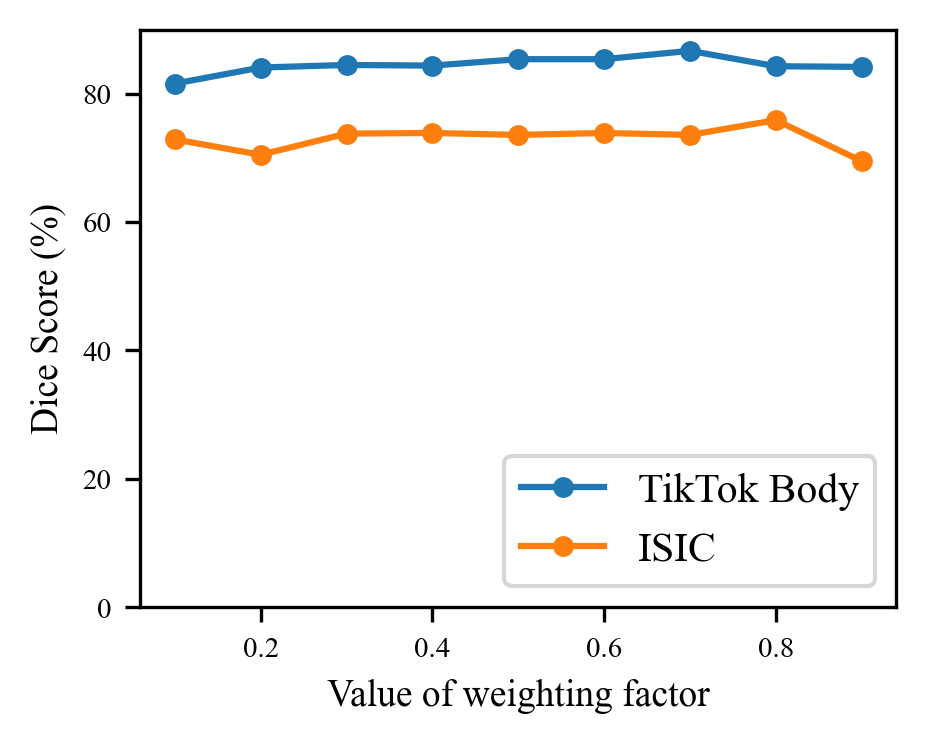} 
\caption{Ablation studies on weighting factor $\alpha$. AM-SAM can perform better with a more carefully selected $\alpha$.}
\label{ablation on alpha}
\end{figure}
\subsection{Ablation Studies}
In this section, we first present ablation studies conducted to assess the individual contributions of the integration of YOLOv8 based object detector for automated prompting and Hadamard Product based mask calibration components to the overall performance of our model. The experiments were performed using ISIC and TikTok Body datasets, and the results are summarized as illustrated in Table~\ref{tab:ablation}.

When trained with 8 examples on TikTok Body dataet, if YOLO-generated bounding box prompts were included in the model (without mask calibration), the performance improved to 85.2\%. This indicates that the automated prompting mechanism provided by YOLOv8 significantly enhances the model's ability to generate more accurate segmentation, thus leading to better performance. Incorporating the mask calibration module (without YOLO) led to a Dice score of 84.8\%. This improvement over the baseline suggests that the mask calibration process, which refines the segmentation masks, contributes positively to the model's accuracy. When both the YOLO-generated prompts and mask calibration were employed together, the model achieved the highest performance, with a Dice score of 86.7\%. This result clearly demonstrates that the combination of these two modules has a synergistic effect, leading to a significant improvement over both the baseline and the individual contributions of each module. We also conducted an ablation study on the weighting factor $\alpha$ in Figure~\ref{ablation on alpha} with different datasets, which shows that our proposed AM-SAM can perform even better with a more carefully selected $\alpha$. 

\section{Conclusion}
In this paper, we introduce a novel approach, AM-SAM, to enhance the performance and efficiency of BLO-SAM by integrating an YOLO based object detector for automated prompting and employing Hadamard Product for mask calibration for semantic segmentation tasks. By leveraging YOLOv8, we are able to generate more precise initial prompts, which not only enhance the model’s segmentation performance but also accelerate the convergence during training. Additionally, the use of Hadamard Product in the mask calibration process further refines the segmentation masks, contributing to more precise and reliable outcomes. Our improved results of experiments with few-shot training set demonstrate the effectiveness and performance of our approach in segmentation tasks.




\bibliography{aaai25}

\clearpage
\appendix
\section{Technical Appendix}
\subsection{Datasets}
In this paper, we evaluate our method mainly on three datasets, including two on the general domain and one on the medical domain, each of which is publicly available. The two general domain datasets are used for human body segmentation task, and the medical domain dataset is for skin lesion segmentation task. The statistics of the test sets are listed in Table~\ref{tab:test sets}.
\begin{table}[H]
\centering
\caption{Number of test examples in different tasks.}
\begin{tabular}{lccc}
\toprule
Dataset & TikTok & Vikram Shenoy & ISIC-2018 \\ 
\midrule
Test size  & 2000 & 100 & 1000 \\ 
\bottomrule
\end{tabular}
\label{tab:test sets}
\end{table}

\noindent
\textbf{(1) Human Segmentation Dataset - TikTok Dances~\cite{anwar2021tiktok}:} For the human body segmentation task, this dataset is a specialized dataset curated for the task of human segmentation within dynamic and visually diverse TikTok dance videos. The dataset comprises 2,615 images extracted from TikTok videos, each meticulously segmented to isolate the dancing figures from their backgrounds. This dataset is particularly challenging due to the high variability in poses, movements, and backgrounds typical of TikTok dance content. For our experiments, we utilized the last 2,000 images from this dataset as the testing set, allowing us to assess the generalization capability and the robustness of our segmentation model.

\noindent
\textbf{(2) Human Segmentation Dataset~\cite{shenoy_human_segmentation_2024}:} The Human Segmentation Dataset hosted on GitHub by Vikram Shenoy provides a collection of images designed for human figure segmentation across various environments. This dataset includes 300 images of people in both indoor and outdoor settings, offering a diverse range of scenarios to train and evaluate segmentation models. We utilized the last 100 images as test set.

\noindent
\textbf{(3) ISIC-2018 Challenge Dataset~\cite{codella2018skin}:}
It is a well-known dataset in the medical imaging community, specifically designed for the task of skin lesion segmentation. The dataset contains dermoscopic images of skin lesions, annotated with pixel-wise labels to delineate areas of interest. It was used in the International Skin Imaging Collaboration (ISIC) 2018 challenge, aimed at advancing research in automated melanoma detection. The dataset is challenging due to the variability in lesion appearance, size, and location, and it is a benchmark for evaluating the performance of segmentation algorithms in medical image analysis. We utilized the training and testing sets provided by the ISIC-2018 Challenge without making any modifications to the original dataset splits as outlined on the official website.
\subsection{Qualitative Results}
\begin{figure}[t]
\centering
\includegraphics[width=1.0\columnwidth]{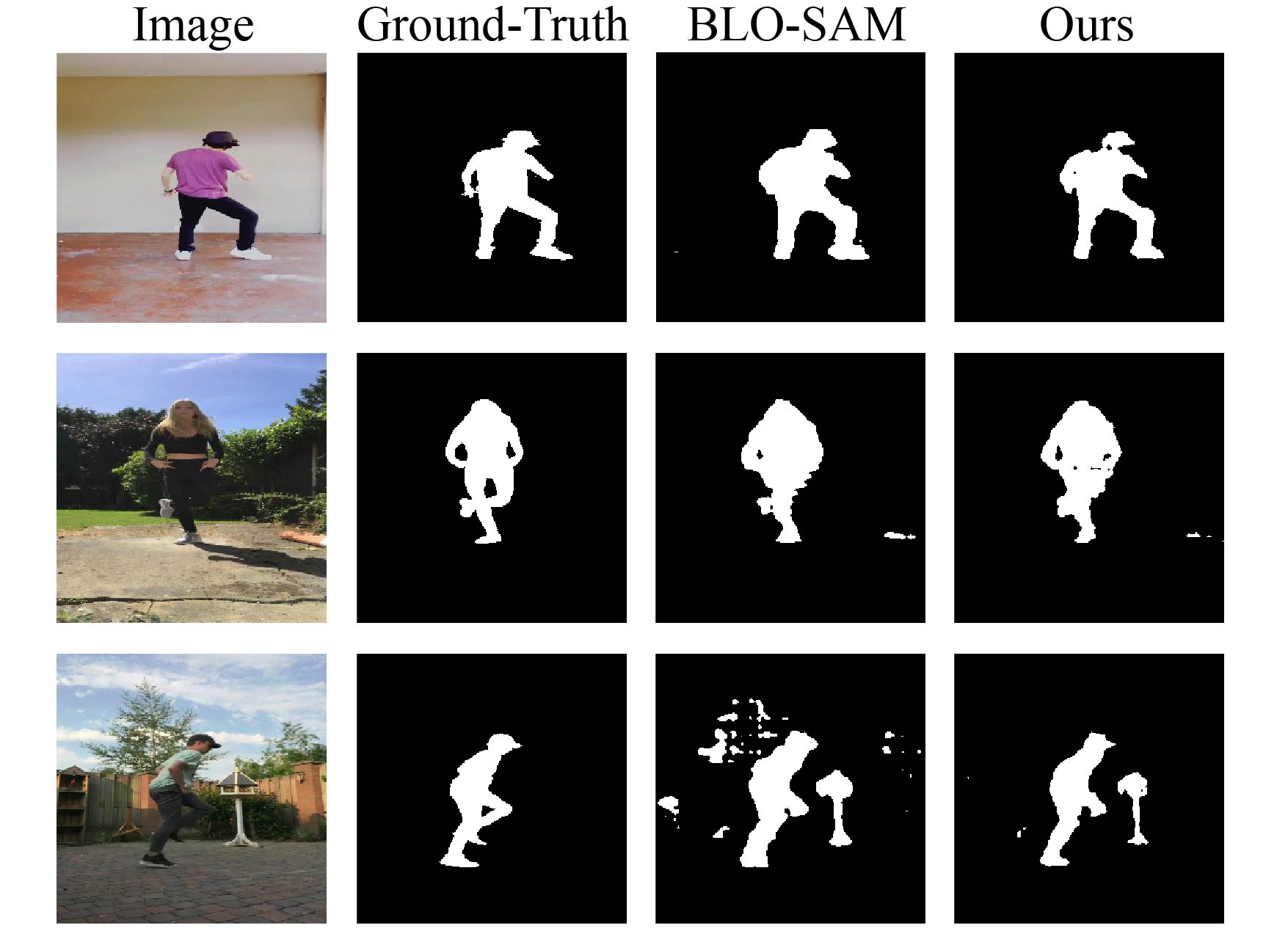} 
\caption{Qualitative results on some randomly sampled test examples from TikTok body dataset.}
\label{tiktok}
\end{figure}

\begin{figure}[t]
\centering
\includegraphics[width=0.9\columnwidth]{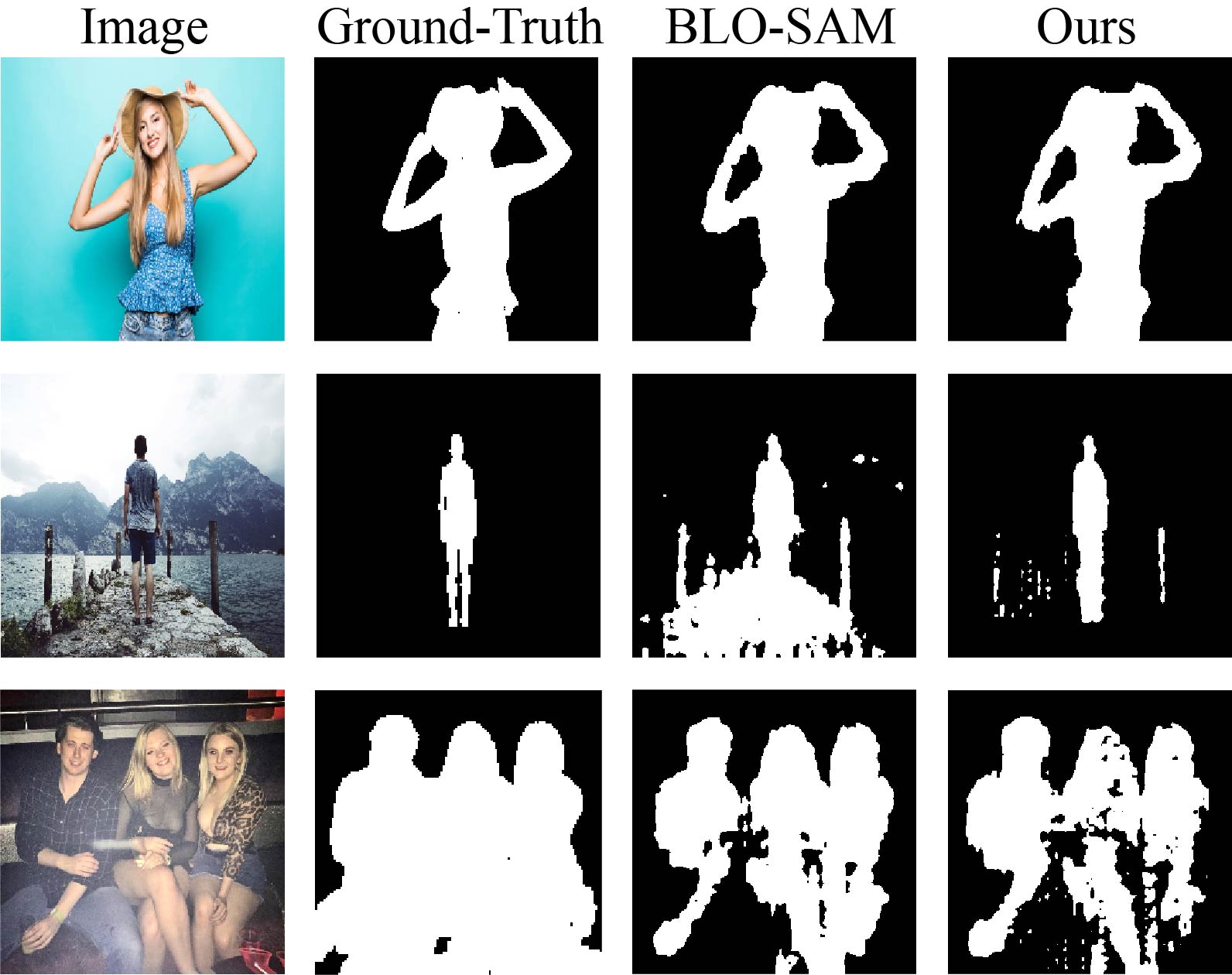} 
\caption{Qualitative results on some randomly sampled test examples from Vikram Shenoy body dataset.}
\label{body2-qual}
\end{figure}

In Figures~\ref{tiktok} and ~\ref{body2-qual}, we present the segmentation masks generated by beseline method BLO-SAM and our proposed AM-SAM approach. Notably, across a range of tasks, AM-SAM consistently outperforms BLO-SAM in terms of the precision and clarity of the segmentation masks. The masks predicted by AM-SAM exhibit an enhanced level of detail, with more accurate delineation of target components and reduced background noise. This improvement is particularly evident when compared to BLO-SAM, which, while effective, tends to include more background artifacts and less precise boundaries. The qualitative results clearly demonstrate that AM-SAM not only captures intricate features with greater fidelity but also maintains a cleaner separation between the foreground and background, highlighting its superior performance across diverse segmentation challenges.

\end{document}